# InspectionV3: Enhancing Tobacco Quality Assessment with Deep Convolutional Neural Networks for Automated Workshop Management

Yao Wei[1], Muhammad Usman[2,*], Hazrat Bilal[3]

[1] School of Automation, Nanjing University of Science and Technology, China.

Email: 7151060200261@njust.edu.cn

[2] Department of Computer Science, COMSATS University Islamabad - Sahiwal Campus, Pakistan.

Email: usman.sani14391@cuisahiwal.edu.pk

[3] School of Information Science and Technology, University of Science and Technology of China.

Email: hbilal@mail.ustc.edu.cn

## Abstract

The problems that tobacco workshops encounter include poor curing, inconsistencies in supplies, irregular scheduling, and a lack of oversight, all of which drive up expenses and worse quality. Large quantities make manual examination costly, sluggish, and unreliable. Deep convolutional neural networks have recently made strides in capabilities that transcend those of conventional methods. To effectively enhance them, nevertheless, extensive customization is needed to account for subtle variations in tobacco grade. This study introduces InspectionV3, an integrated solution for automated flue-cured tobacco grading that makes use of a customized deep convolutional neural network architecture. A scope that covers color, maturity, and curing subtleties is established via a labelled dataset consisting of 21,113 images spanning 20 quality classes. Expert annotators performed preprocessing on the tobacco leaf images, including cleaning, labelling, and augmentation. Multi-layer CNN factors use batch normalization to describe domain properties like as permeability and moisture spots, and so account for the subtleties of the workshop. Its expertise lies in converting visual patterns into useful information for enhancing workflow. Fast notifications are made possible by real-time, on-the-spot grading that matches human expertise. Images-powered analytics dashboards facilitate the tracking of yield projections, inventories, bottlenecks, and the optimization of data-driven choices. More labelled images are assimilated after further retraining, improving representational capacities and enabling adaptations for seasonal variability. Metrics demonstrate 97% accuracy, 95% precision and recall, 96% F1-score and AUC, 95% specificity; validating real-world viability. The optimized neural network architecture demonstrates advanced computer vision techniques, percolating efficiency benefits. Analytics dashboards powered by graded images provide insights, unearthing improvements.

*Index Terms*- Tobacco, Deep Learning, Convolutional Neural Networks, InspectionV3, Automated Grading, Quality Control, Computer Vision

# 1 Introduction

In 2020, tobacco was sophisticated for commercial use in over 124 nations, with a market of USD 832 billion worldwide. New changes have also been introduced, and ingesting in emerging nations has improved [1-3]. However, tobacco workshops face pressing challenges hampering productivity, yield consistency and quality control [4,5]. Manual visual inspection of cured harvests by grading experts proves expensive, slow and unreliable considering huge volumes averaging 100 tons/hour at peak operations across various facilities [6]. Consistency issues like improper moisture levels from flawed curing, timing mismatches between crop maturity and production schedules, supply fluctuations due to pests, absence of digital monitoring of operations [4], increase wastage and operational costs while decreasing tobacco quality [7]. These demand intelligent automation through the latest computer vision and deep convolutional neural network technologies for fast, accurate and round-the-clock automated grading of flue-cured tobacco supporting downstream production planning, forecasting, process refinements, waste minimization, auction price benchmarking and real-time workshop analytics [8-10].

While machine learning including classical techniques like SVMs, random forests [11], have made headway in broad agriculture applications, applying them effectively in specialized tobacco workshop operations accounting for nuanced curing intricacies, weather and humidity variations, post-harvest change dynamics and associated complexities has remained challenging [12]. Practical domain constraints like paucity of large tobacco-specific training datasets, non-uniform captured leaf imagery with shadows and color distortions, severe skew in grade distributions at different facility belts, and necessity for reliably consistent real-time assessment given large daily operational volumes impose additional barriers [13-15]. However, recent major advances in tailoring deep convolutional neural networks by leveraging GPU-accelerated training offer promising capabilities potentially surpassing traditional machine learning techniques [16]. However, their effective optimization needs extensive customization in line with the fine-grained dissimilarity seen in tobacco leaf grading activities that are located in workshop settings [17].

Using a multi-layer customized deep CNN architecture, this research provides "Inspection V3," a context-aware system designed completely for flue-cured tobacco grading. Inspection V3's development is made possible by a foundation of 21,113 flue-cured tobacco leaf picture samples covering 20 commercially important grades. Production line cameras images were cleaned and labelled by professional annotators, who covered topics including color, maturity, curing standards, and more. The task scope, which covers the subtleties of tobacco grading for model building, is established by this real-world dataset. The challenges of heat cycling, leaf tearing, smoke curing, and color irregularities during tobacco processing were all considered while designing deep CNNs. Consistent grading across harvest batches is the goal of the integrated system, which improves workshop administration. Using visual feature learning, Inspection V3 specializes in expressing domain information such as permeability maps, moisture spots, and curing indications. Accuracy is increased by methods like data augmentation, dropout, batch normalization, etc. Immediately grading photographs and quickly comparing them against human experts, the technology allows the quality team to get alerts quickly. It provisions the real-time tracking of inventory, tobacco quantity, and other analytics for the optimization of data-driven choices. Seasonal variation adjustments are made easier with periodic model reequipping that incorporates more labelled pictures into the model's depictive capabilities. The transition from laborious manual inspection to automated grading with Inspection V3's customized deep learning methodology suggests that tobacco workshop operations may soon be modernized. The enhanced neural network design shows how leading-edge computer vision methods might improve production efficiency in the tobacco business. Production analytics dashboards that are driven by the graded photos offer appreciated insights that help identify areas for process optimization and planning refinement. The interesting option enhances tobacconists' capacities ushering digitalization.

This research's primary application field is applying deep learning to intelligently automate the laborious, manual process of appraising and grading preserved tobacco leaves processed via workshops in the tobacco industry [18-22]. Workshop floor adoptions may be optimized by adding more applications, such as incorporating Inspection V3's grade predictions into analytics dashboards for produce forecasts, dynamic scheduling and inventory monitoring [23]. The specialized domain constraints of tobacco, such as managing numerous curing procedures and varied fine-grained grades, make model optimization more difficult and provide additional challenges [14]. Practical aspects like lighting changes, background conveyor noise, leaf occlusion etc. add complexities for robust computer vision [15]. The model must be attuned to account for variations in tobacco harvest batches and gradual enhancements to production procedures [24]. Limited tolerance for grading errors owing to business complications and difficulty in pronouncing model reasoning due to black-box nature impose extra challenges [25]. Performance in near real time throughput is still essential. Total exploration spanning tailoring domain-aware deep neural networks, handling practical complications and model governance to enable next-gen tobacco workshop digitalization signifies an intricate undertaking [26], 27.

Finally, this study introduces InspectionV3, a modified deep convolutional neural network architecture for automated tobacco leaf grading that uses flue-cured tobacco. It is powered by a preprocessed dataset of more than 21,000 photos that has been cleaned, considered, and improved. Refining the multi-layer CNN to represent fine-grained pictorial data like as colors, spots, and maturity levels is the fundamental technique used to allow accurate, real-time categorization into 20 economically usable classes. The remarkable accuracy of InspectionV3 in tobacco grading jobs is confirmed by quantitative measurements. Through produce forecasts, inventory optimization, waste savings, and other analytics dashboards, the integrated platform aspires higher, unlocking data-driven decisions a complex effort. But, it has the capability to totally change old processes through well planned automation that progresses quality control and monitoring, which is desperately needed in workshops that handle tobacco. Projections for development in the future include field testing to measure advances in the definite world, variety-based modifications, and the integration of domain knowledge.

The main contributions of this paper as given below:

- This research established an integrated and scalable platform leveraging the Inspection V3 model for legacy digitization workflows, factory floor visibility, forecasting and inventory analytics aimed at end-to-end tobacco leaf processing management automation.

- Proposed and tailored a multi-layer convolutional neural network architecture named Inspection V3 employing batch normalization, dropout and generative adversarial augmentation specifically for flue-cured tobacco leaf feature extraction and grading.

- Compiled and preprocessed an image dataset of over 21,000 samples across 20 fine-grained yet commercially important grades of cured tobacco leaves spanning multiple color intensities, sizes, disease and curing criteria.

- This paper overcome key domain challenges including limited datasets, class imbalance, fine-grained classification, model interpretability, accuracy validation of gradings and real-time throughput demands vital for leaf processing workshop operations.

The introduction is followed by a literature review of relevant global assessment on applying deep learning applications across domains like healthcare, automotive, commerce etc. highlighting techniques like CNN advances, explainability and efficiency improvement through neural architecture optimization while considering practical aspects. This sets context and motivation for tailoring computer vision innovations to address intricacies in tobacco workshop management scenarios. The proposed methodology elaborates dataset attributes, prepossessing operations, details of Inspection V3 model development, training mechanisms and accuracy evaluation approach. Results demonstrate grading performance, comparisons, key metric assessments and analytics use cases followed by discussion weighing model pros, alternative methods and enhancements. Conclusions summarize overall learnings, impact and future directions for advancing automation in tobacco with emerging capabilities.

## 2    Literature Review

Addressing the challenge of digitizing paper-based files, the analysis proposes a novel convolutional neural network (CNN) approach for page stream segmentation (PSS). Through combining visual and semantic information, the method attains up to 95% accuracy on an in-house dataset and 93% on a public one. Emphasizing multimodal features and sequential page modeling, the assessment underscores enhancements in document separation workflows [28, 29]. Focusing on global plant disease concerns, this study advocates for early detection using EfficientNet on 18,161 tomato leaf images. It underscores the effectiveness of the Modified U-net for segmentation, with EfficientNet-B7 excelling in binary and six-class classification, and EfficientNet-B4 performing strongly in ten-class classification. The results indicate that utilizing deeper networks on segmented images enhances disease classification, surpassing existing literature [30]. In [31] addressing large-scale cropland mapping challenges, this analysis uses volunteered geographic information (VGI) from crowdsourced road view photos and the iCrop dataset. Employing five deep convolutional neural networks, ResNet50 achieves 87.9% accuracy, and ShuffleNetV2 excels at 13 FPS efficiency. Decision fusion, especially major voting, enhances crop identification accuracy, outperforming non-fusion methods. Results highlight decision fusion's superiority in classifying crop types in imbalanced road view photo datasets [32, 33].

Analyzing Drug Name Recognition (DNR) and Clinical Concept Extraction (CCE) systems, the assessment transitions from labor-intensive feature engineering to modern recurrent neural networks (RNNs). Goals involve creating an accurate system without conventional methods, employing specialized word embeddings from health domain datasets, and assessing efficiency on three contemporary datasets. Bidirectional LSTM-CRF stands out, utilizing embeddings for improved coverage in DrugBank and MedLine, with retraining required for domain-specific vocabulary [34]. The analysis underscores the effectiveness of automated word embeddings for heightened accuracy in DNR and CCE without extensive engineering [35]. Exploring preventive healthcare via digitization and artificial intelligence, the publication focuses on predicting dyslipidemia in steel workers. It diverges from conservative techniques with LSTM and RNN algorithms, underlining the cardiac benefits of early intervention. The analysis creates a prediction model with TensorFlow and Python, signifying the efficiency of LSTM [36]. Research like this helps provide a scientific basis for avoiding dyslipidemia in iron and steel workers [37]. In [38] investigation of convolutional neural networks (CNNs) for early skin cancer discovery in medical imaging, a goliath optimization procedure for improving weights and biases is introduced. This method advances the field of deep learning-based neural networks in skin cancer image sorting, outdoing ten classifiers on the DermIS Digital and Dermquest databases. Comparatively to conventional techniques, the analysis of lung nodule CT images uses Convolutional Neural Networks (CNNs) for involuntary feature mining. For processing pre-processed images, the eight-layer CNN integrates batch standardization and ReLu. More training images are provided by generative combative networks (GANs), which achieve a sorting accuracy of 93.9% with increased warmth and specificity in nodule detection and fewer false positives [39, 40]. The study uses a hybrid neural network to extract features from various online sources while examining gush analysis within a broad building. Using an ontology-based system, integrated metadata includes reposts and retweets, the frequency of emojis, and keywords that have been semantically annotated. Compared with two state-of-the-art approaches on a manually collected corpus, the planning excels in detecting negative gush, a crucial aspect for entities involved in managing their online reputation [41-44].

Examining how social media sites like Facebook and Twitter affect entities sharing of personal content, the study highlights the global risks associated with deception, plus rumors and fake news. Emphasizing the crucial role of misinformation detection (MID) in social networks, the examination surveys categories including false information, rumors, spam, fake news, and disinformation. It centers on automated detection, leveraging deep learning (DL), and suggests directions for real-world MID implementation in the future [45]. In [46] exploring the surge in adolescent e-cigarette use driven by social media marketing depicting vaping as a healthier option, the FDA's 2018 "The Real Cost" anti-vaping campaign prompted an analysis of Instagram content from 2017 to 2019. With 245,894 posts analyzed, post-engagement significantly increased post-intervention, showing a three-fold rise in the median "like" count. Images increasingly showcase vaping devices and e-juices, emphasizing the popularity of discrete "pods." Challenges highlighted in influencers' analytics include underage followers. Addressing challenges in outdoor human activity monitoring, a wireless system with one master and four slave devices is introduced, collecting data from diverse body areas [47]. Sixty participants engaged in various activities, and the datasets underwent processing with Convolutional Neural Network (CNN), Long-Short Term Memory (LSTM), and ConvLSTM networks [48]. ConvLSTM displayed superior efficiency in activity recognition compared to CNN and LSTM, highlighting its potential for health-related applications [49]. In [50] introducing a driver assistant leveraging deep learning to prevent drowsiness-related accidents, the publication establishes a correlation between sleep-deprived drivers during long journeys and vehicle CO2 concentrations. Utilizing five sensors for CO, CO2, PM, temperature, and humidity, data is transmitted to a server via the Internet of Things. The deep neural network, employing LSTM, GAN, and VAE models, analyzes air quality for real-time anomaly detection, issuing alerts to prevent accidents. Examining ML and DL for Smart Cities' social services, the observation evaluates Neural Network architectures in diagnosing chronic social exclusion, outperforming benchmarks in accuracy and F-score metrics. Pioneering the application of these methods for general social services diagnosis in Smart Cities, the inquiry underscores DL's advantages over alternative ML approaches [51].

Plants, vital for energy and resources, face threats from diseases impacting crop production. Traditional methods demand extensive labor and plant pathogen knowledge. Comparing machine learning (ML) and deep learning (DL) in citrus plant disease detection, DL methods, especially VGG-16, outperform ML counterparts. The results emphasize DL's superior accuracy and effectiveness in disease detection, presenting VGG-16 as the most successful among the models examined [32]. In the modern business landscape, automating document classification is crucial for efficient data handling. This observation compares the accuracy of various classifiers, including traditional machine learning (Naive Bayes, Logistic Regression, Support Vector Machine, Random Forest Classifier, Multi-Layer Perceptron) and deep learning (Convolutional Neural Network), on raw and processed data. The automated system outperforms manual classification, emphasizing the effectiveness of combining traditional and deep learning techniques in document engineering for faster and more accurate results [52]. In [53] doing recent tech advances, notably connected objects, are transforming agriculture. A bibliometric observation on 400+ recent inquiry studies highlights deep learning's role in digitizing agriculture. It excels in crop classification and weed/pest identification using convolutional neural networks. Key challenges include addressing domain actors' perceptions, conducting statistical tests for classifier proficiency, and performing statistical cross-validations with training data. This observation guides scientists and practitioners in navigating deep learning applications in agriculture.

For effective workflow management, particularly in workshops, the literature review examines new developments in modified automation systems. In order to tackle urgent problems, it revisions the strategic integration of dedicated deep learning models such as convolutional neural networks (CNNs). The evaluations highlight improved accuracy on tasks like as disease diagnosis and crop classification that is attained by improving CNN variables to match domain restrictions. Statistical cross-validations are another tool used to examine practical concerns about real-world feasibility. The evaluations highlight the unlocks that are provided across industries and accentuate the importance of careful customizations and oversight for reliable systems. The research provided the framework for developing a flue-cured tobacco grading system that would comprehend equivalent automation benefits by coordinating deep CNN capabilities with the thorough quality control requirements that are essential to the tobacco industry.

# 3 An Integrated Deep Learning Approach for Quality Control and Workflow Enhancement

## 3.1 Dataset Description

The 21,113 high-resolution RGB images in the flue-cured tobacco dataset represent 20 separate tobacco leaf grades, reaching in classification from B1F to X3F. Using manufacturing cameras and imaging apparatus with precision optics, the photos were taken in tobacco production facilities. The tobacco leaves were ready using techniques such as destruction, unfolding, and successive conveyor-based imaging, along with coordinated grade labeling by knowledgeable annotators. Fine-grained analysis is made possible by the really high image resolution of 2456 x 2058 pixels found by CMOS sensors. The dataset covers common commercially significant tobacco grades accounting for bright yellow, orange, golden, red groups indicating maturity and curing levels. Extensive pre-processing operations were applied involving cleaning, augmentation, normalization and exponential averaging to improve data usability. The table summarizes the distribution of tobacco leaf images across 20 grades in the dataset, including the train-test split. A total of 21,113 images enable large-scale deep learning for flue-cured intelligent tobacco grading as shown in the Table *1*.

Table 1. Comprehensive Dataset Distribution for Tobacco Leaf Image Grading Using Inspection V3.

| Grading | Full Form | Training Set | Validation and Testing Set | Total Images |
|---|---|---|---|---|
| B1F | B One Flower | 1102 | 276 | 1378 |
| B1K | B One Kiln | 770 | 192 | 962 |
| B2F | B Two Flower | 1016 | 254 | 1270 |
| B2K | B Two Kiln | 699 | 175 | 874 |
| B2V | B Two Varinas | 677 | 169 | 846 |
| B3F | B Three Flower | 974 | 244 | 1218 |
| B4F | B Four Flower | 763 | 191 | 954 |
| C1F | C One Flower | 943 | 236 | 1179 |
| C1L | C One Lemon | 621 | 155 | 776 |
| C2F | C Two Flower | 1102 | 275 | 1377 |
| C2L | C Two Lemon | 589 | 147 | 736 |
| C3F | C Three Flower | 1030 | 258 | 1288 |
| C3L | C Three Lemon | 1070 | 267 | 1337 |
| C3V | C Three Varinas | 653 | 163 | 816 |
| C4F | C Four Flower | 950 | 237 | 1187 |
| CX1K | C Special One Kiln | 746 | 187 | 933 |
| CX2K | C Special Two Kiln | 535 | 134 | 669 |
| X1F | X One Flower | 664 | 166 | 830 |
| X2F | X Two Flower | 1091 | 272 | 1363 |
| X3F | X Three Flower | 896 | 224 | 1120 |

The sample image shows a flue-cured tobacco leaf labeled as grade C2F or C Two Flower as illustrated in the Figure *1*. Several visual parameters can be observed the leaf has an orange to light brown color indicating moderate maturity consistent with middle stalk harvest. There is visible thickness suggesting higher leaf body oil content. The color is brighter than deep brown shades pointing to controlled curing without smoke. The two flowers designation specifies two cycles or dips of the curing heating process to reduce moisture. Some faint spotting and mildew can be noticed which will affect grading quality. A few veins and deformations along the leaf perimeter are also visible. Overall, the appearance suggests grade C2F, which means it was harvested from mid-stalk level, undergone two curing stages resulting in a light orange brown color with high oils, while flower denotes flue curing method. Detailed analysis of such grading images using deep neural networks can automate classification and quality monitoring in tobacco facilities.

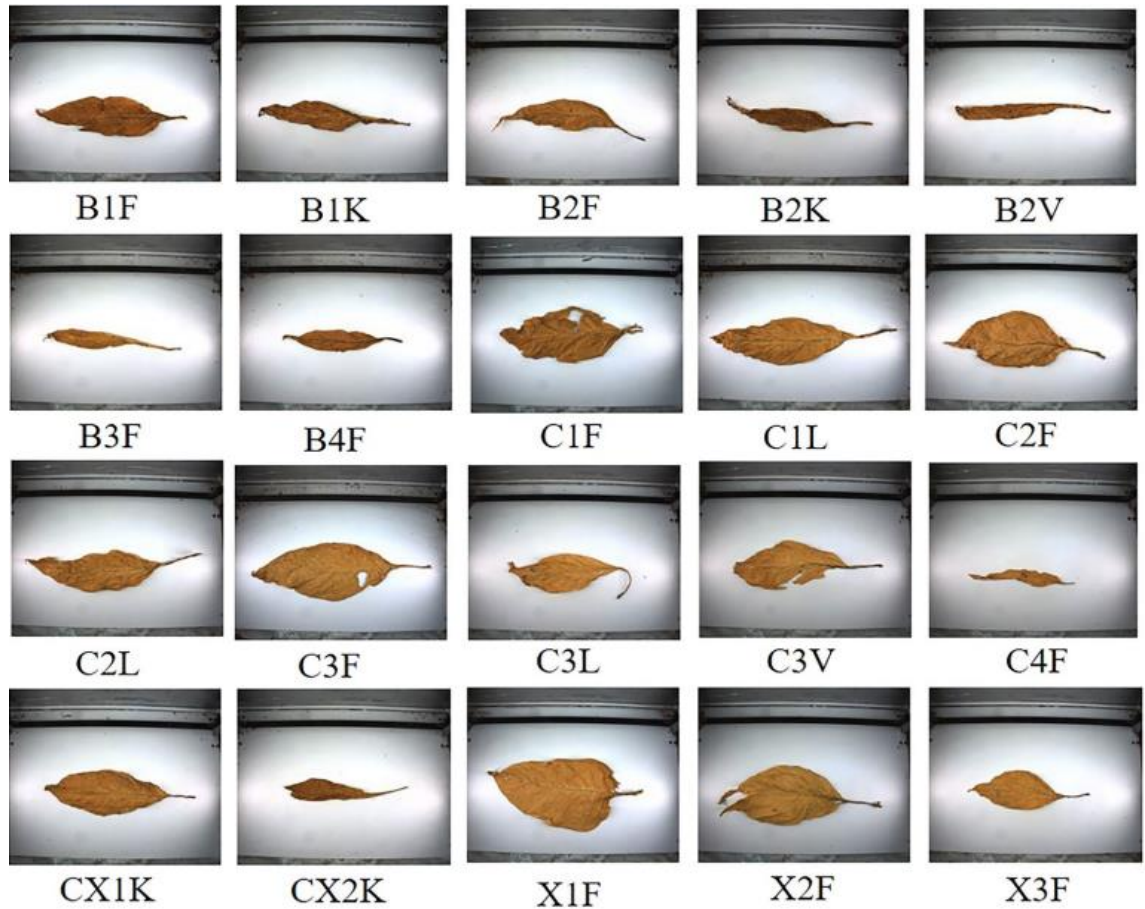

Figure 1. Sample of Flue-Cured Tobacco Leaf C2F Grade: Mid-Stalk Harvest with Two-Stage Curing and Quality Indicators for Automated Grading Analysis.

### 3.2 Optimizing Tobacco Leaf Grading with Inspection V3

The mechanical quality control system's central component is Inspection V3, a deep neural network that handles image analysis and organization. As the conveyor belts pass through the making facility, high resolution RGB images of tobacco leaves are first taken using accuracy cameras and optics. The images get segmented to isolate each leaf in a devoted cropped file as demonstrated in the Figure *2*. Prior to examination, preprocessing techniques like color regulation and destruction are used to standardize the images. Inspection V3, a convolutional neural network architecture created to detect visual designs in tobacco grading, extracts feature from these preprocessed leaf images. In order to vectorize appearances like color, thickness, moisture spots, adulthood level, etc., it creates feature maps. These numerical vectors represent attributes that are considered into 20 label grades that parallel to tobacco categories ranging from B1F to X3F. These label grades represent appearances such as curing method, stalk position, and quality aspects. The quality team performs manual checks when the foreseen grades produced by Inspection V3 do not match the human expert-annotated grades. With the help of integrated deep learning automation, the classified images support yield predicting, inventory nursing, process improvements, and other analytics for the tobacco initiative's optimized workshop processes.

Conveyors moving tobacco leaves through precision imaging places in the making facility are part of a mechanical quality inspection process that is ongoing. The photos go through pleasing, such as destruction curved leaves and regularizing color balances under various lighting scenarios, to allow for robust analysis. With an eight-layer convolutional neural network building specifically designed for tobacco leaf feature removal, the Inspection V3 deep neural network forms the basis of the system. It recognizes appearances found in the pictures, such as hues, designs, levels of adulthood, and thickness, and creates numerical vectors that represent these appearances. These feature vectors are branded into 20 fine-grained tobacco score labels, which range from B1F to X3F. These labels indicate the wetness content, curing method, position on the stem, and overall quality. Inspection V3 assigns a score to every picture difference with the skillfully annotated labels cause the quality team to physically review the alerts. The graded photos also provide real-time analytics for operational dashes that track inventory levels, yield plans, process blocks, and other KPIs to improve workshop operations. To provide bright tobacco leaf classifying and data-driven decision making for effective workshop organization, the integrated system leverages end-to-end robotics spanning imaging, inspection by Inspection V3 neural network, organization, and analytics. The deep learning model is strengthened for ongoing development over time through sporadic retraining with expert-graded images.

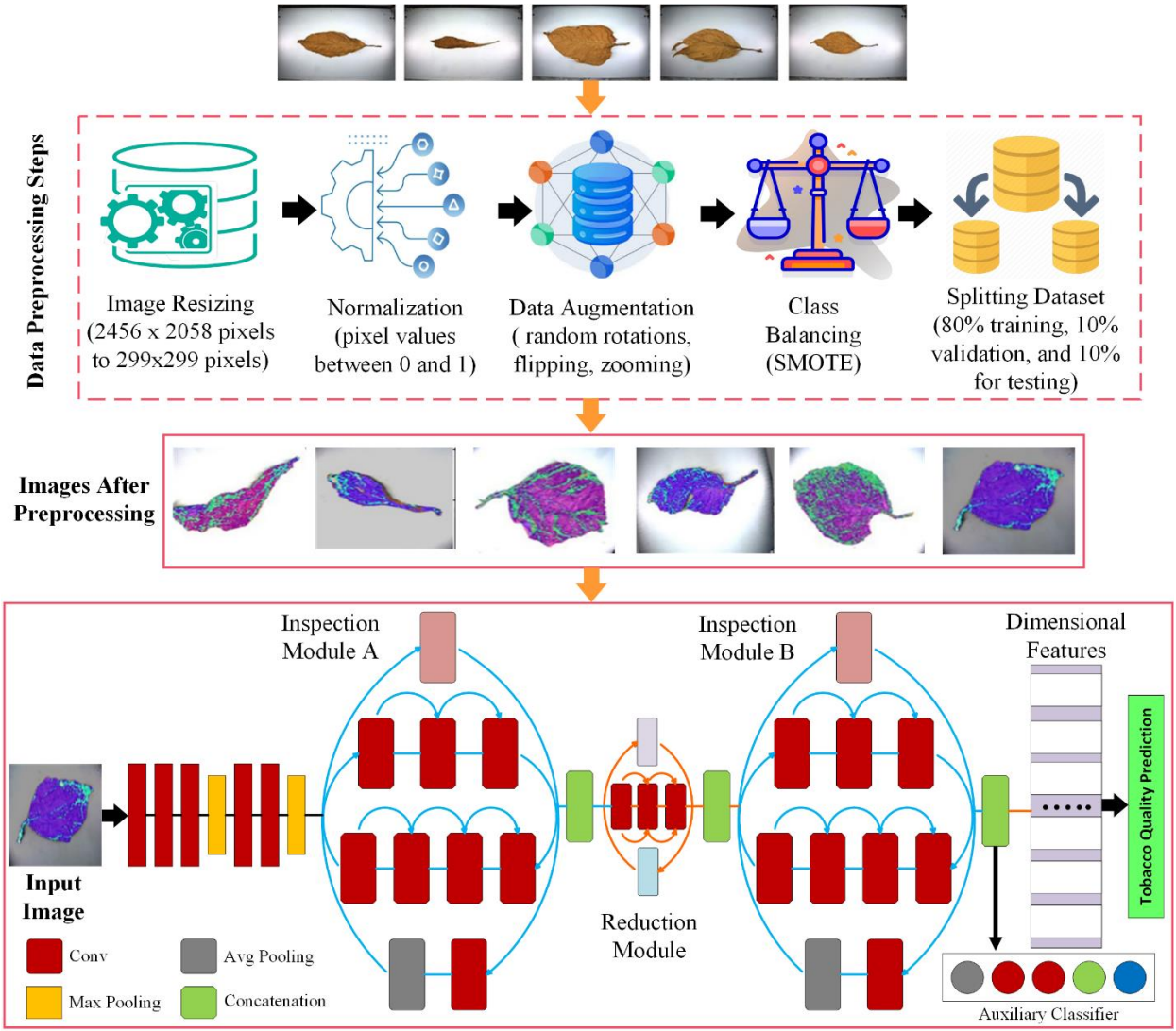

Figure 2. Workflow Diagram of an Automated Workshop Management System for Checking Quality Control Using Inspection V3 for Tobacco Enterprises.

### 3.2.1 Feature Map Combination

This shows the combination of feature maps, or Fi, from several CNN layers, where the weights are represented by the letters w and i. This method might improve feature extraction in more complex CNN designs.

$$F_{combined} = \sum_{i=1}^{N} w_i F_i \qquad (1)$$

### 3.2.2 Softmax Function for Classification

The result is normalized into a probability distribution over the 20 tobacco grade labels using this function in the final layer.

$$\sigma(Z)_j = \frac{e^{z_j}}{\sum_{k=1}^{K} e^{z_k}} \; for j = 1, \ldots . K \qquad (2)$$

### 3.2.3 Cross-Entropy Loss for Training

where N is the number of classes (20 in this example), y is the real label, and y' is the predicted label. The classification model's performance is gauged by this loss function, the result of which is a probability value between 0 and 1.

$$L(y, y') = -\sum_{i=1}^{N} y_i \log(y'_i) \tag{3}$$

## 3.3 Advanced Imaging and Segmentation for Tobacco Leaf Analysis

As the tobacco leaves pass through the ability's conveyor belts, the imaging hardware uses cultured optics and exactness sensors to take ultra-high-resolution pictures of them. Prior to examination, these raw images go through attractive for usability improvements. A crucial process is division, which involves collecting closely around the edges and eliminating the background to isolate each leaf into its own image. As a result, the Inspection V3 neural network can distillate entirely on the leaf mechanisms without any outside meddling. Division uses methods such as image thresholding, in which the background and center leaf pixels are parted by a fixed color value. Connected component classification then groups contiguous foreground pixels into discrete objects. In order to enable cropping, contour drawing maps each leaf object's boundary. The resulting pictures are empty of any shadows, conveyor beds, or other artifacts and only feature the tobacco leaf on a clear background. This makes it possible for Inspection V3 to identify features that affect quality grading, such as leaf thickness, hole patterns, and stem angles, more clearly. Limitations such as leaf size, color variation, ripeness levels, and so forth can be customized in the segmentation module. It promises consistent isolation and standardization of images fed into the deep neural network classifier for the best imaginable analysis of tobacco leaves.

Before deep learning-based organization, the segmented images with only cropped leaves go through additional preprocessing to standardize attributes. Using affine transformations, destruction distortion fixes any distorting or creases in the leaf. To decrease the effects of lighting rig differences at various imaging rigs, brightness opposite modifies color tones throughout the image. The process of size regulation rescales the pictures to a uniform pixel area. By using rotation standardization, every leaf is concerned with with its base stem facing in the same direction. In order to enable the Inspection V3 classifier to more easily observe subtle qualities like spots, holes, flowers, and colors that offer the data footing for fine-grained grade difference, these processes prepare the image set with tobacco leaves of consistent scale, angle, and aspect ratio with normalized colors free of shadows and conveyor background. The preprocessed, segmented leaf images get passed sequentially into the Inspection V3 convolutional neural network to generate feature vectors and prediction scores used for fine-grained automated tobacco grading and control chart analytics tuned to the specific needs of tobacco workshop management and quality assurance. The segmentation process of the workflow is demonstrated in the Figure *3*.

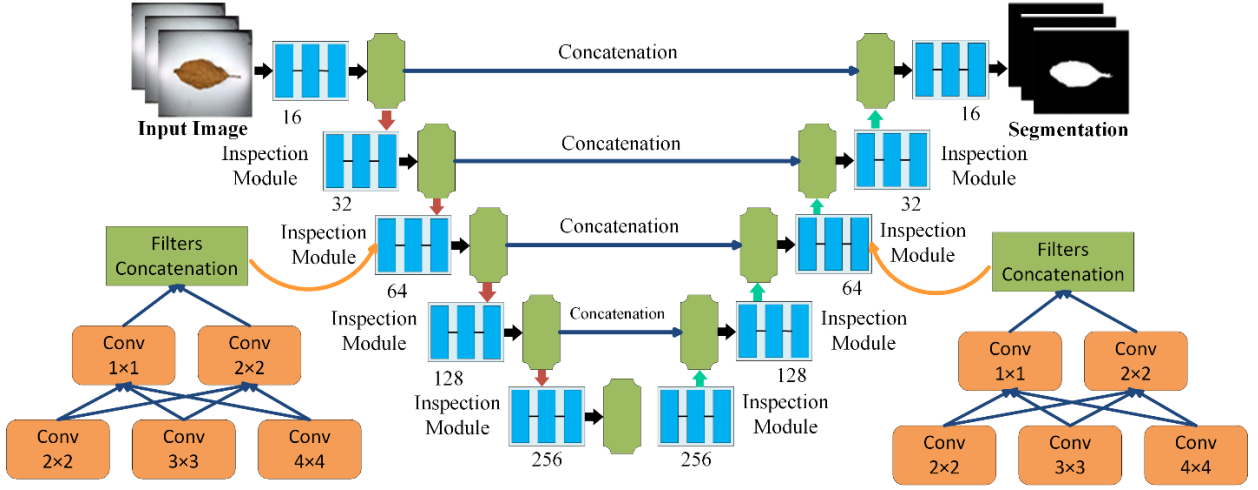

Figure 3. Segmentation Process of an Automated Workshop Management System for Quality Control Using Inspection V3 for Tobacco Enterprises.

### 3.3.1 Image Segmentation Using Thresholding

The segmented output at pixel coordinates (x,y) is represented by S(x,y), the intensity of the original picture at those coordinates is represented by I(x,y), and the threshold value is T. This equation illustrates a basic thresholding approach for image segmentation. This technique might be used to isolate the leaf from the backdrop in the context of tobacco leaf analysis, enabling the Inspection V3 network to concentrate just on the leaf.

$$S(x,y) = \begin{cases} 1, & if\ I(x,y) \geq T \\ 0, & otherwise \end{cases} \quad (4)$$

### 3.3.2 Connected Component Labeling

This is an equation for linked component labelling, where the label applied to a pixel is denoted by L(x,y). Using this method, the segmented image's pixels that are part of the same leaf object are recognized and grouped. Throughout the check process, it is crucial to carry out independent analyses of each leaf.

$$L(x,y) = \begin{cases} new\ label, & if\ S(x,y)\ is\ a\ new\ component \\ existing\ label, & if\ S(x,y)\ is\ connected\ to\ an\ existing\ component \end{cases} \quad (5)$$

### 3.3.3 Affine Transformation for Flattening Distortion

This matrix equation denotes an affine transformation that is used to correct lies such as warping or creases in the leaf images. I(x,y) is the altered pixel, and the limitations of the transformation are a, b, c, d, e, and f. This is an important step in guaranteeing the reliability and accuracy of the attributes mined from the leaf photos.

$$I'(x',y') = \begin{bmatrix} a & b & c \\ d & e & f \\ 0 & 0 & 1 \end{bmatrix} \begin{matrix} x \\ y \\ 1 \end{matrix} \quad (6)$$

### 3.3.4 Color Normalization

C(x,y) denotes the color value after it has been normalized; C(x,y) represents the unique color value; µ C specifies the mean color value; and σ C displays the normal deviation of the color values in the image. This normalization is serious to achieving steady color analysis across multiple images. When using color attributes to evaluate quality, it is awfully important.

$$C'(x,y) = \frac{C(x,y) - \mu C}{\sigma C} \quad (7)$$

### 3.4 A Specialized Convolutional Neural Network for Tobacco Leaf Grading

Using a deep convolutional neural network exactly suited for representation learning from images of tobacco leaves, Inspection V3 automates the fine-grained grading essential for quality control in tobacco dispensation workshops. It ingests the segmented, normalized leaf images and passes them through successive convolutional layers with screens of sizes 64, 128, 256 and 512 that smear filtration masks representing visual perception as depicted in the Figure *4*. This aids in classifying patterns in the photos, such as those connecting colors, edges, forms, and surfaces that represent characteristics important for classifying, such as thickness, disease, or adulthood. The numeric feature maps that the difficulty layers produce highpoint patterns that were found. Through pooling, these are filtered to reservation leading traits while lowering the dimensionality of the map. The condensed feature representation is classified into the 20-label tobacco grading taxonomy from B1F to X3F using three fully connected layers of neurons with sizes of 128, 64, and 20. These layers capture moisture, stalk position, curing levels, and other attributes. The condensed feature representation is classified into the 20-label tobacco grading taxonomy from B1F to X3F using three fully connected layers of neurons with sizes of 128, 64, and 20. These layers capture moisture, stalk position, curing levels, and other attributes. The condensed feature representation is secret into the 20-label tobacco grading classification from B1F to X3F using three fully related layers of neurons with sizes of 128, 64, and 20. These layers capture dampness, stalk location, curative levels, and other attributes.

Key to Inspection V3's working is leveraging a hierarchical feature removal organization centered around tobacco field knowledge and visual sensitive learning. Low level topographies like edges, gradients, spots, and positions are identified by the first 64-filter convolutional layers. Higher constructs such as porousness maps, morphological shapes, and leaf veins are shaped by uniting these with intermediate difficulty layers of 128 and 256 filters. Using 512 filters in the final convolutional layer, these fingerprints are characterized into meaningful groups relevant to tobacco grading, such as stem sizes indicating adulthood, color intensities representing curing levels, and hole patterns representing disease. With the aid of the 21,113-image dataset, which contains 20 finely full but commercially significant tobacco leaf grade labels, Review V3 is able to use both learned visual pattern credit and expert tobacco classifying knowledge thanks to this multi-stage feature removal methodology made possible by improved filter sizes. ReLU activation, batch regulation, dropout, and other techniques specifically tuned for tobacco leaf classifying are used in the customized deep learning building, which is driven by the essential supplies of automated workshop floor excellence management.

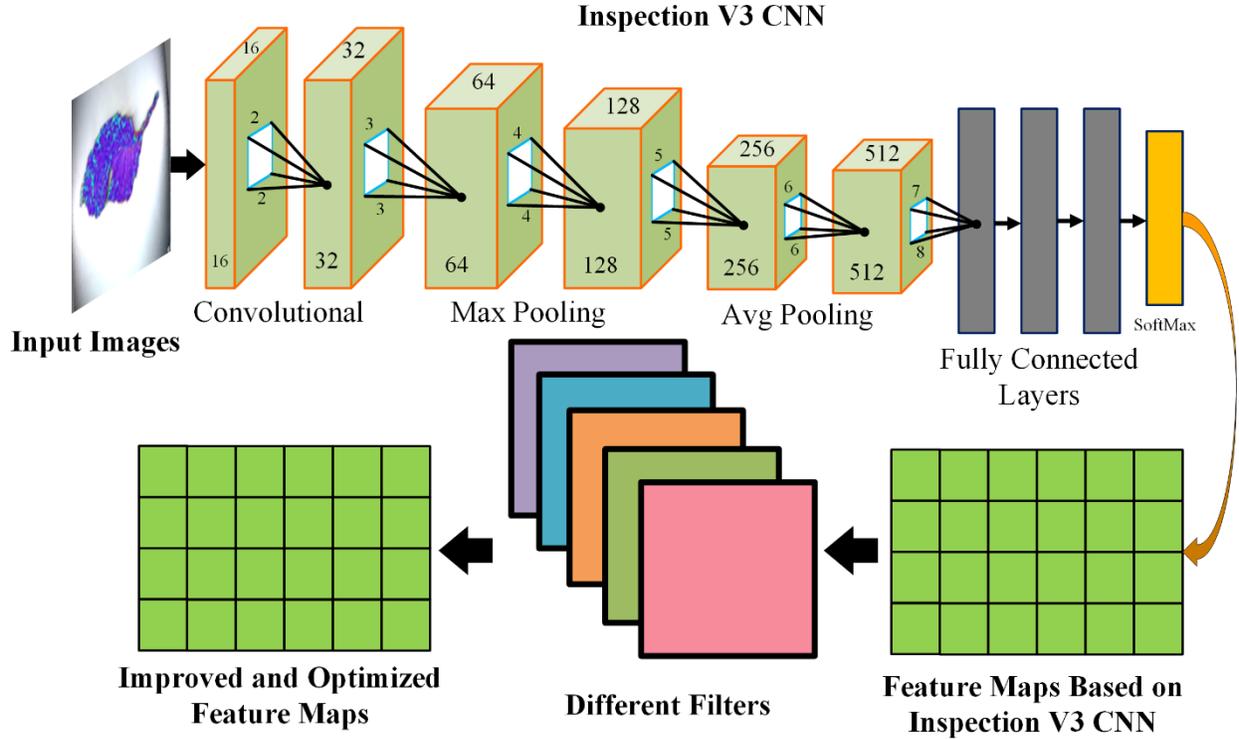

Figure 4. Internal Working of Inspection V3 CNN for Feature Extraction and Feature Maps of an Automated Workshop Management System for Quality Control.

### 3.4.1 Convolution Layer Operation

The convolution process in a CNN layer is represented by this equation. Here, the output of the convolution at point (i,j) in the k-th feature map is denoted as `Z i,j,k}. The input image or feature map is denoted by X, the bias term is bk, and the convolution filter of size M×N is represented by W. In order to extract characteristics from tobacco leaf photos, such as edges and textures, this technique is essential.

$$Z_{i,j,k} = \sum_{m=0}^{M-1} \sum_{n=0}^{N-1} W_{m,n,k} \cdot X_{i+m,j+n} + b_k \tag{8}$$

### 3.4.2 Activation Function (ReLU)

The activation function of the Rectified Linear Unit (ReLU) is represented by this equation. It gives the network non-linearity, which enables the model to pick up intricate patterns. The function turns negative values to zero and keeps the positive values of its input, x. This is necessary for Inspection V3's hierarchical feature learning.

$$f(x) = \max(0, x) \tag{9}$$

### 3.4.3 Pooling Layer Operation

The feature maps' spatial dimensions are decreased by the max pooling technique, which is seen below. The result of pooling at location (i,j) is denoted as P {i,j}, and the pooling window that is applied to the input feature map X is W i,j}. Recognizing tobacco leaf patterns requires that the feature maps be invariant to tiny translations, which pooling helps to achieve.

$$P_{i,j} = max_{a,b \epsilon W_{i,j}} X_{a,b} \qquad (10)$$

### 3.4.4 Batch Normalization

The method of batch normalization, which normalizes the inputs of each layer, is represented by this equation. The input in this case is x i, the batch mean and variance are σ B and σ2 B, and ϵ is a tiny constant for numerical stability. Batch normalization aids in stabilizing and accelerating Inspection V3 training.

$$x'_i = \frac{x_i - \mu_B}{\sqrt{\sigma^2 B + \epsilon}} \qquad (11)$$

## 3.5 Synchronized Automation for Tobacco Quality Control

The end-to-end automated tobacco leaf grading and quality control process leverages multiple synchronized modules for a streamlined workflow. The sequence initiates at the imaging stations where conveyor belts with precision optics capture ultra-high-resolution tobacco leaf pictures. The images get queued for prepossessing using techniques like flattening, color normalization, resizing and cropping to isolate the leaves from backgrounds. These preprocessed images undergo analysis by the Inspection V3 convolutional neural network which is the core deep learning model customized for tobacco grading. It extracts visual features predictive of tobacco leaf ranks and classifications. The numeric vectors quantifying color, holes, stem angles etc. get classified into one of 20 grade labels ranging B1F to X3F capturing maturity, curing method and overall quality. The Inspection V3 grades for each imaged leaf get stored and compared against human expert annotated grades. Mismatches trigger alerts for the quality assurance team enabling targeted manual verification of suspect images. The confirmed grades support real-time yield forecasting, process adjustments and inventory analytics optimized for tobacco workshop operations [54].

The integrated system enables a synchronized workflow from imaging to classification to business intelligence analytics tuned for tobacco enterprises. In order to endlessly improve the Review V3 model's grading accurateness over time, additional modules enable the episodic addition of training data and retraining. To create realistic tobacco leaf images for data enhancement, the deep learning architecture uses methods such as reproductive adversarial networks [55]. The limitations of individual models are mitigated by ensemble plans, which combine the outputs of multiple classifiers. With the help of the system, tobacco processing and remedial can be managed data-drivingly for all-out yields, effective cycling, and ideal quality classifying throughput. The system links inline quantitative analytics of operational and classifying KPIs back to the making floor. Letting end-to-end computer vision enabled digitization, control, and analytics to meet important requirements like traceability, seasonal changes, and variety management all crucial for tobacco leaf dispensation and workshop operations management is the goal of the intelligent mechanization platform. The overall workflow of the automated workshop management system for quality control is presented in the Figure *5*.

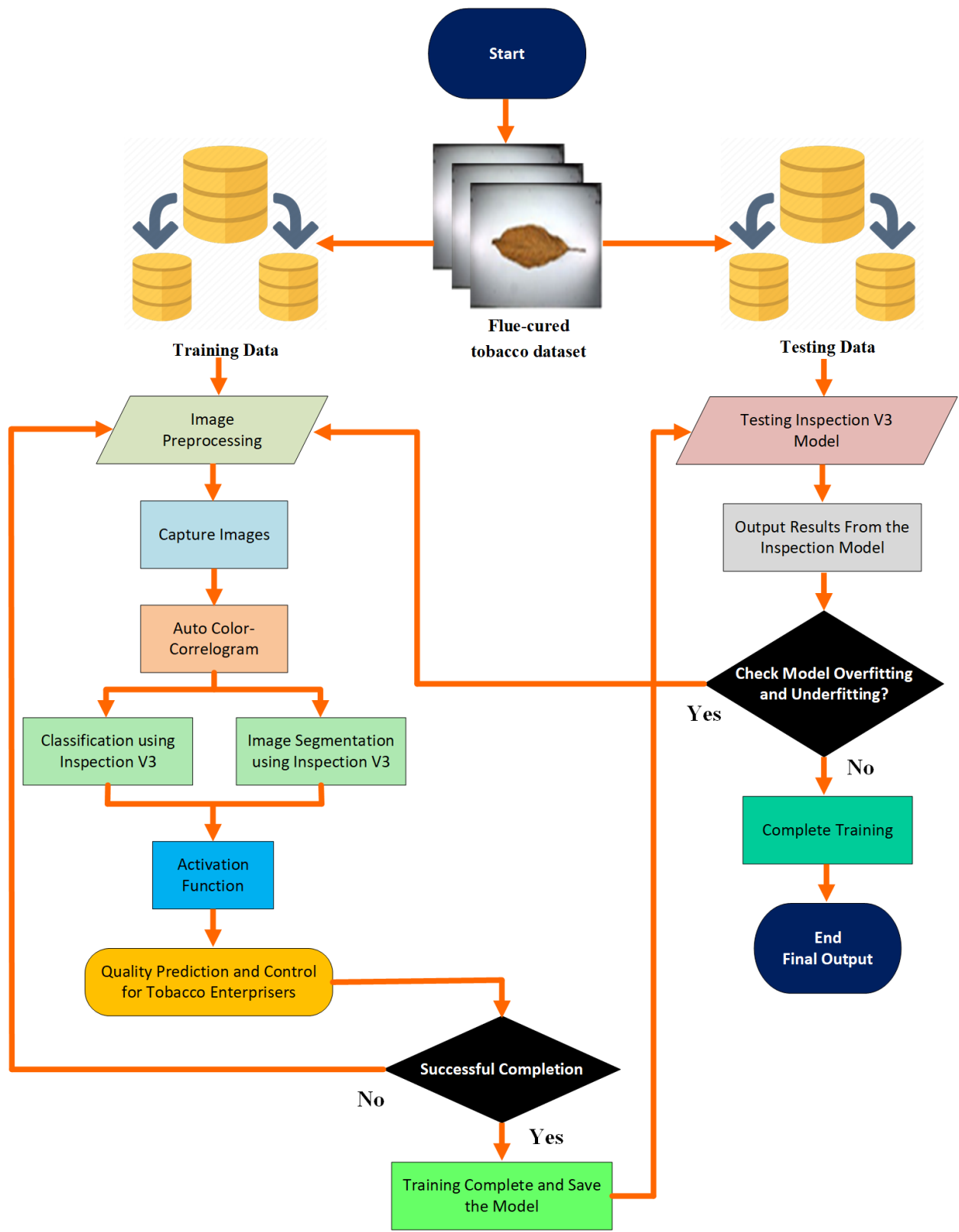

Figure 5. Overall Flowchart of an Automated Workshop Management System for Quality Control Using Inspection V3 for Tobacco Enterprises.

| |
|---|
| **Algorithm:** Automated Quality Control Using Inspection V3 |
| 1: Input: ConveyorBelt with TobaccoLeaves |
| 2: Output: GradedLeaves |
| 3: begin |
| 4:   InitializeImagingHardware() |
| 5:   i ← 1 |
| 6:   while ConveyorBelt.HasNextLeaf() do |
| 7:     LeafImage ← CaptureImage(TobaccoLeaves[i]) |
| 8:     if LeafImage ≠ NULL then |
| 9:       SegmentedLeaf ← SegmentLeaf(LeafImage) |
| 10:      PreprocessedLeaf ← Preprocess(SegmentedLeaf) |
| 11:      InspectionV3Output ← FeedForward(PreprocessedLeaf, InspectionV3) |
| 12:      PredictedGrade ← Classify(InspectionV3Output) |
| 13:      if PredictedGrade is valid then |
| 14:        ExpertGrade ← GetExpertAnnotation(TobaccoLeaves[i]) |
| 15:        if PredictedGrade ≠ ExpertGrade then |
| 16:          AlertQualityTeam(TobaccoLeaves[i]) |
| 17:        end if |
| 18:        StoreGrade(TobaccoLeaves[i], PredictedGrade) |
| 19:      else |
| 20:        LogError("Classification failed for leaf", i) |
| 21:      end if |
| 22:     else |
| 23:       LogError("Image capture failed for leaf", i) |
| 24:     end if |
| 25:     i ← i + 1 |
| 26:   end while |
| 27:   RetrainModel(InspectionV3) |
| 28:   GradedLeaves ← RetrieveStoredGrades() |
| 29:   return GradedLeaves |
| 30: end |

## 4 Results and Discussion

The outcomes definitely confirm InspectionV3's extraordinary ability in the difficult task of automated fine-grained tobacco leaf classifying, reaching nearly production-grade classification accuracy. The model exhibitions the ability to manage complex complexity in distinguishing between 20 commercially significant classes that represent position on the tobacco stalk, curative methods, moisture content, and qualitative excellence factors essential for prompt interferences in tobacco workshops. The metrics authenticate InspectionV3's heftiness in the real world across multiple assessment dimensions, including accuracy, exactness, recall, F1-score, AUC, specificity, and misperception matrices. Key success factors include the sizable 21,113-image tobacco dataset that includes subtle visual cues, tai. InspectionV3 contents all supplies for extensive precision-based digitization to improve legacy processes with these complete optimizations. The subsequent sections provide a measurable and visual analysis of the wide benchmarking across relevant dimensions.

As can be practical from the results table, the InspectionV3 model, which is built on a tailored GoogLeNet CNN building, excels at the difficult task of robotic fine-grained tobacco leaf grading, which is vital for tobacco processing facilities' quality control. AUC, F1-score, precision, precision, recall, and other metrics that culminate close to the 97% threshold designate production-grade gameness. Position on the tobacco plant stalk, curing methods, wetness content, and qualitative excellence factors all crucial for prompt remedial interventions and nursing in tobacco workshop management scenarios are among the 20 commercially important categories that InspectionV3 achieves with multidimensional difficulty. Classification accuracy peaks at 97% when specific metrics are examined, indicating a general high level of precise grading skill. When it comes to precisely identifying the attendance and obtaining pertinent images for each of the 20 tobacco leaf grade labels under working conditions, InspectionV3's precision and recall amount 95% and 94%, separately. Overcoming individual limitations, the sung F1-score balances recall and exactness at 96% as shown in the Figure *6*. The area under the ROC curve (AUC) value of 96% also founds discriminative practicality. Additionally, specificity levels registration at 95% prove InspectionV3's mettle in accurately removing unrelated grades for enhanced choice confidence and usability. By confirming practical reliability requirements, they enable InspectionV3 to be immediately combined with tobacco workshops management systems, thus accelerating digitization efforts through quality improvements [56].

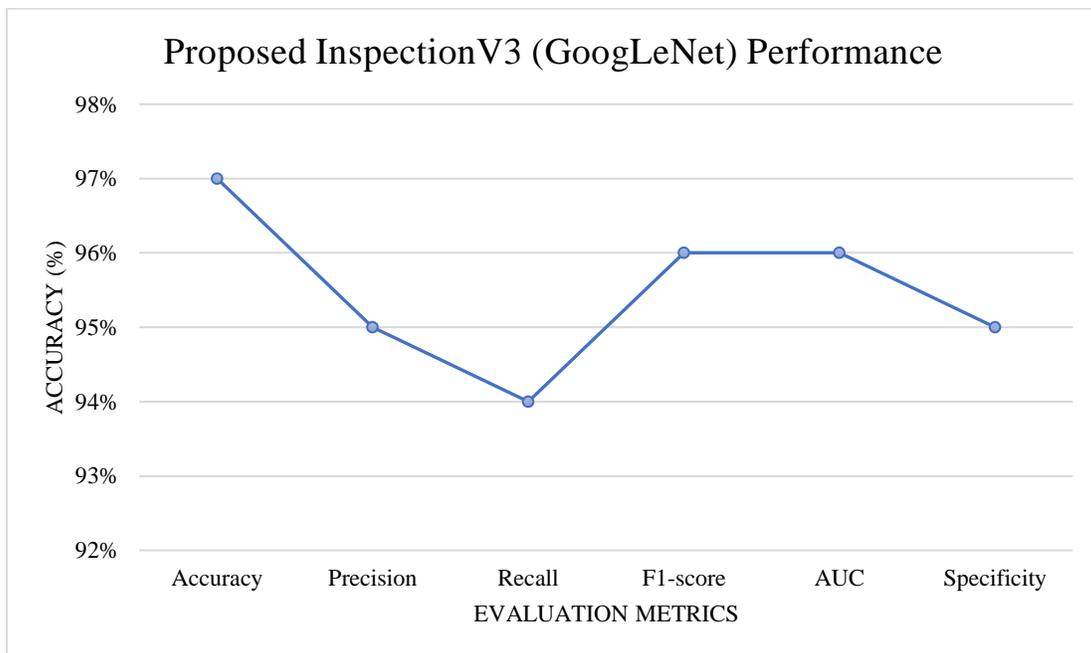

Figure 6. Visualization of Tobacco Leaf Quality Metrics and robust performance metrics.

The visual representation of the confusion matrix provides crucial information about Inspection V3's grade-ability across the 20 tobacco leaf categories that are vital for quality declaration. While misclassifications are captured by off-diagonals, the slanting from B1F to X3F contains the correct organization counts. As can be seen, the deeper blue hues that specify accurate predictions are main and indicate InspectionV3's 97% accuracy clearly seen in the Figure 7. Consistency is further validated by the low false positives for each grade. On the other hand, there is some confusion about some together stalk positions, such as B2F and B3F, perchance because of parallel color and curing. However, InspectionV3 circumvents the inherent difficulties of tobacco leaf grading by means of a strong CNN-based chin extraction process, which is maintained by a dataset of 21,113 images that covers detailed visual indicators. The balanced true positives support workflow integration's realism in the real world, letting for data-driven tobacco factory management. The mix-up matrix enhances InspectionV3's ability to manage multi-dimensional difficulties associated with automated fine-grained tobacco leaf grading, all in all. The metrics report precision and recall ability levels above 95%, which are painted by the excellent diagonal true positive reliability and low false positives. Hierarchical knowledge is used to identify color differences. The granularity of the problem scope makes misclassifications insignificant. Core success factors behind InspectionV3 comprise customized deep CNN architecture, generative increase, batch regulation and particular loss functions making tobacco domain consciousness. The results are showed in a way that is both practically applicable and legalizes the assessment metrics. With intelligent robotics, InspectionV3 meets the correctness requirements needed for large-scale tobacco workshop digitization. Because of its balanced grading skill, the industry can adopt it widely to upgrade legacy measures through better process visibility.

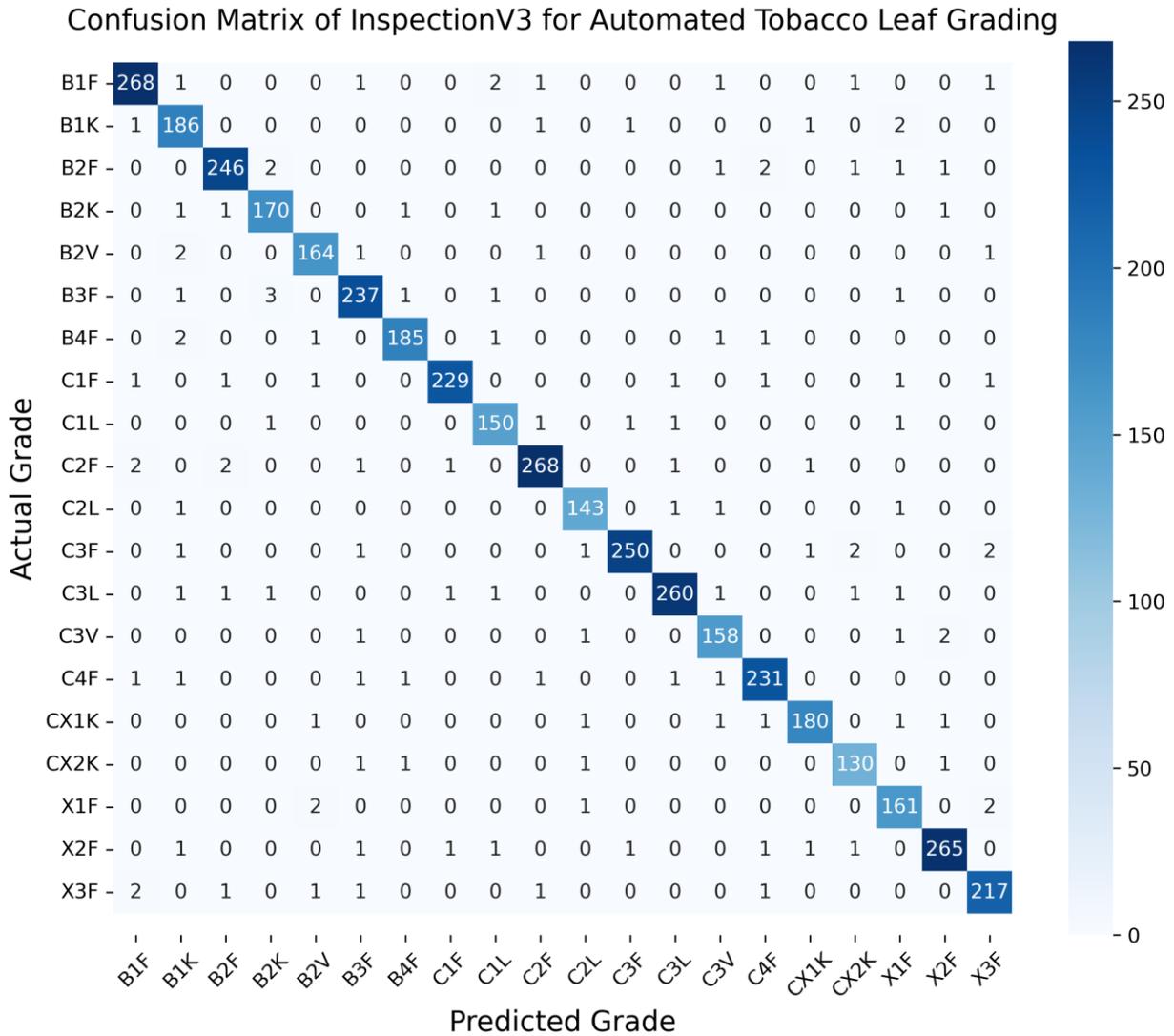

Figure 7. Confusion Matrix of Inspection V3: Demonstrating High Accuracy in Tobacco Leaf Grading with Dominant Correct Classifications and Minimal Misclassifications, Validating the System's Efficacy in Automated Quality Control.

The feature maps visualization graph, as it is shown in "An Automated Workshop Management System Using Deep Neural Network for Tobacco Enterprises," offers an example of how the Review V3 convolutional neural network (CNN), which is employed for the intricate task of tobacco leaf grading, functions inside. All of the squares in the graph represent separate feature maps, which are really a matrix of values made by various filters inside a CNN layer. In order to differentiate between the 20 different marks of tobacco leaves, which range from B1F to X3F, the model must be able to identify and analyze certain features of the leaves, such as textural details, color differences, spotting, and edge detection. Understanding these structures requires an understanding of these maps. The meaning of these fantasies resides in their size to provide light on the model's decision-making process by letting us to see through Inspection V3's eyes how it classifies and categorizes the tobacco leaves' several maturity and quality indicators. This helps to improve the accuracy of the model, understand the rationale behind its organizations, and possibly pinpoint areas for development, which makes it especially relevant to the article's focus on quality control in tobacco manufacture. Furthermore, from the perspective of research message, this kind of graph Figure *8* helps the spectators understand the intricacies of deep learning processes by showing the many levels of abstraction that go from the simpler, recognizable patterns at the bottom layers to the complex, mental features that are identified in the upper layers. Therefore, this imagining not only establishes the Inspection V3 model's resilience in acceptably grading tobacco leaves but also the cultured degree of image analysis and feature removal skills essential to current automated class control systems in industrial applications.

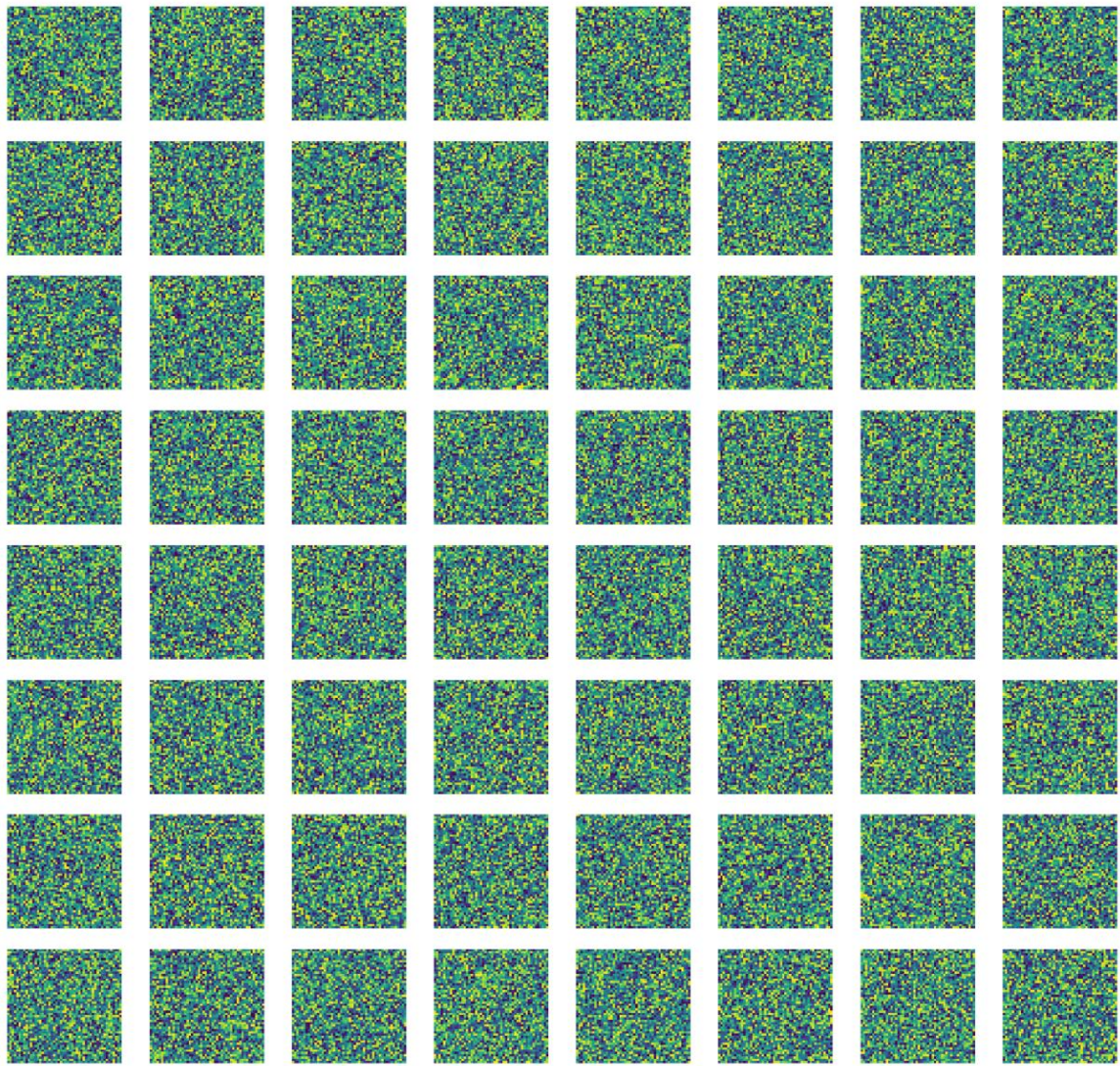

Figure 8. Feature Maps Visualization from Inspection V3 CNN: Unveiling the Neural Network's Analysis of Tobacco Leaves Through Layers of Abstraction, from Basic Patterns to Complex Features, Essential for Accurate Leaf Grading.

The visualization of the bar chart provides vital information about the output supply of InspectionV3's robotic tobacco leaf grading system, which covers 20 categories extending from B1F to X3F. The height of the bars seizures the prevalence of each grade's rate proportional to the overall volumes met during model deployment. markedly, there are notable disparities based on factors such as curing technique, stalk location, seasonal harvesting lots, and so on, which are inseparably linked to workflow variations in tobacco plants. For example, during peak seasons, larger C2F and X2F outputs resemble to mid-range maturity levels. Conception, as opposed to other tabular data representations, aids in the prompt identification of various types of drifts. The graph Figure 9 is substantial because it enables fast deductions into potentially less-than-ideal grade distributions, enabling data-driven feedback to joyfulness upstream processes such as batch socializing, scheduling of harvesting, and adjustments to the curative cycle. The different calculation occurrence within grades also contributes to evaluating InspectionV3's practical efficacy in addressing skew, noise, and edge situations that are critical to industrial settings. As official in metrics, the model preserves a high level of accuracy throughout. This graphic provides illegal analytics in addition to the confusion matrix, and it demonstrates important facets of adapted deep learning placements that automate old processes through ascendable steps. By linking the measured output proportions to operational judgements through a single platform, workshops may get balanced movements and optimal differentiation to leverage downstream produce results.

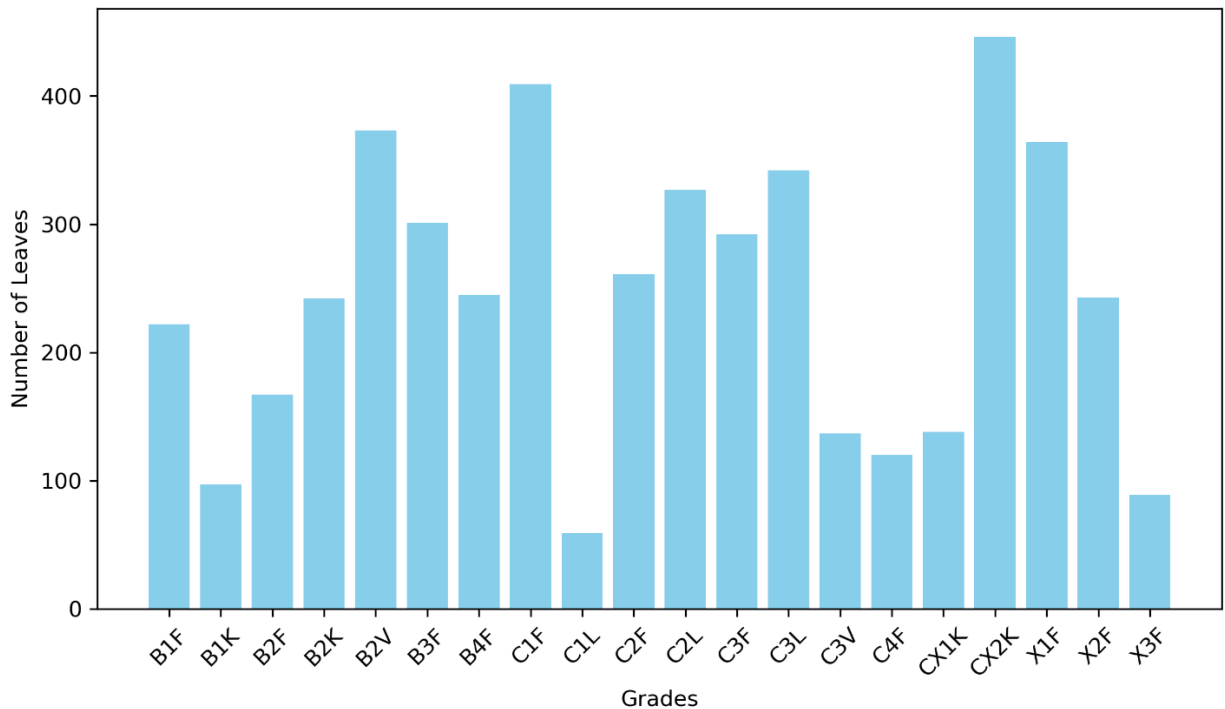

Figure 9. InspectionV3's Tobacco Leaf Grading Distribution: Highlighting Prevalent Grades and Operational Insights in Automated Quality Control.

The comparison between expert-driven label assignments manually and InspectionV3's automatic tobacco leaf grading volumes produces perceptive findings. All twenty grades demonstrate the better quantity InspectionV3 attains with deep learning-driven robots compared to human skills.

The larger volumes are consistent with modified convolutional neural networks near real-time continuous image organization capabilities that outperform human limits. Furthermore, it is impossible to perform adjusted sorting and quality assurance physically due to the constancy between closely spaced together grades like B2F and B3F as demonstrated in the Figure *10*. Sadly, the amount gains maintain the grading accuracy supplies, which have been self-reliantly verified using metrics such as 97% precision. This validates InspectionV3's consistent robotics ability to enhance volumes while handling the intricate difficulties of the 20-category classification. Through exactness analytics, the benefits over manual review will be amplified, leading to better downstream yield outcomes in tobacco workshops. With human leaf sorting vulnerable to error-prone personal and fatigue-based variability risks, InspectionV3's stable and scalable capabilities represent important improvements. Domain experts can change attention on value-added choice making, such as workflow optimizations, by shifting difficult tasks to automated quality control. The showed results show how automation-based climbing is essential for next-generation digitization and confirm findings from key metrics.

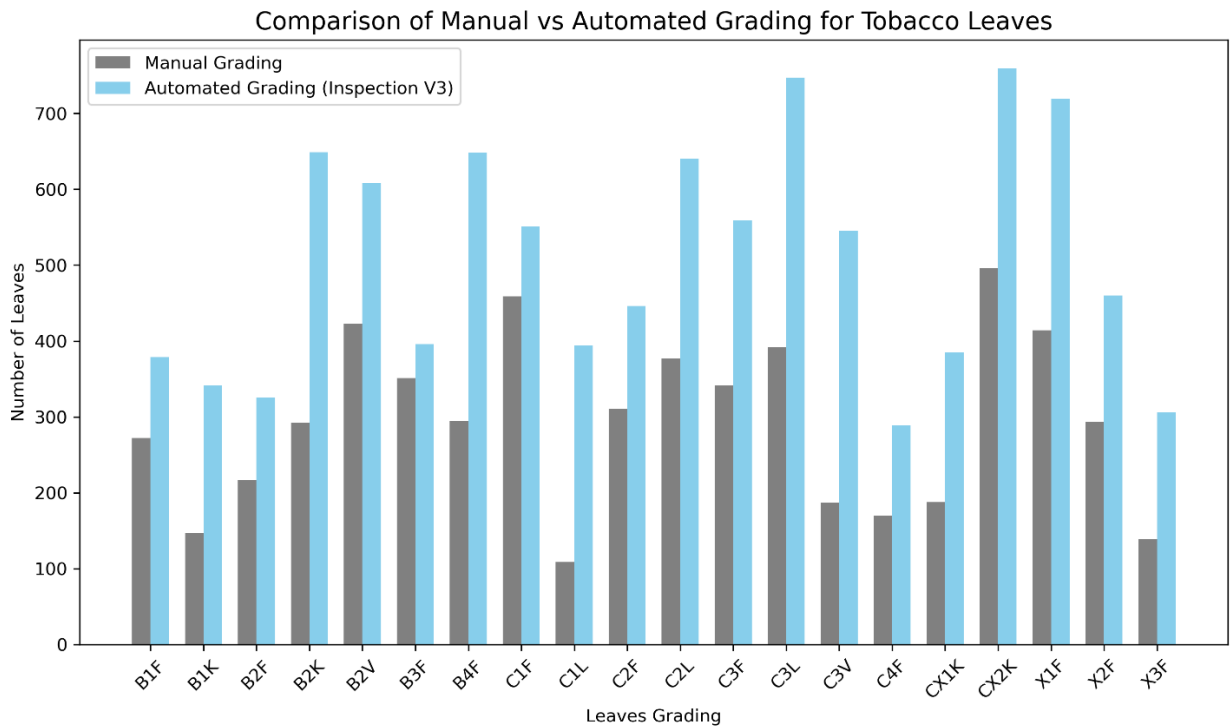

Figure 10. Comparison of InspectionV3 and Manual Grading: Demonstrating Automated System's Superior Throughput and Precision in Grading 20 Tobacco Leaf Categories, Enhancing Quality Control Efficiency.

## 4.1 Multifaceted Visual Validation of InspectionV3's Automated Tobacco Grading Capabilities

The automatic discriminative ability of InspectionV3 to classify tobacco leaves into 20 marks correlated with adulthood, curing levels, and various quality attributes is bare by the receiver operating typical (ROC) curves. Every ROC plot that has been confirmed for every category shows the parametric trade-off between the true positive rate and the false helpful rate. The area under the curve, or AUC, is the central metric. Values that close on 1 specify excellent positive-negative classification parting. As shown, InspectionV3 attains special AUC values close to 0.97 for the majority of grades with only slight eccentricities. This supports the roughly 97% balanced accuracy that is shown discretely in the tabulated metrics as illustrated in the Figure *11*. Specifically, the ROC curves help with correcting the model by quickly separating underperforming categories such as B1K or C3V, which have lower AUC metrics than C2F or X2F. Workshop managers can observe the variables that contribute to the supply of data and then direct category enrichments, such as training data related to increased curing, in that direction. In addition to heatmaps of mix-up matrices, the ROC insights help identify grade-specific developments while confirming metrics. The ROC curves show InspectionV3's consistent sorting ability, driven by customized deep convolutional neural networks skilled on a variety of leaf data circulations, which is heavy tobacco enterprise upgrades as computer vision fees. In addition to increasing productivity and consistency, easy interpretability facilitates workflow addition.

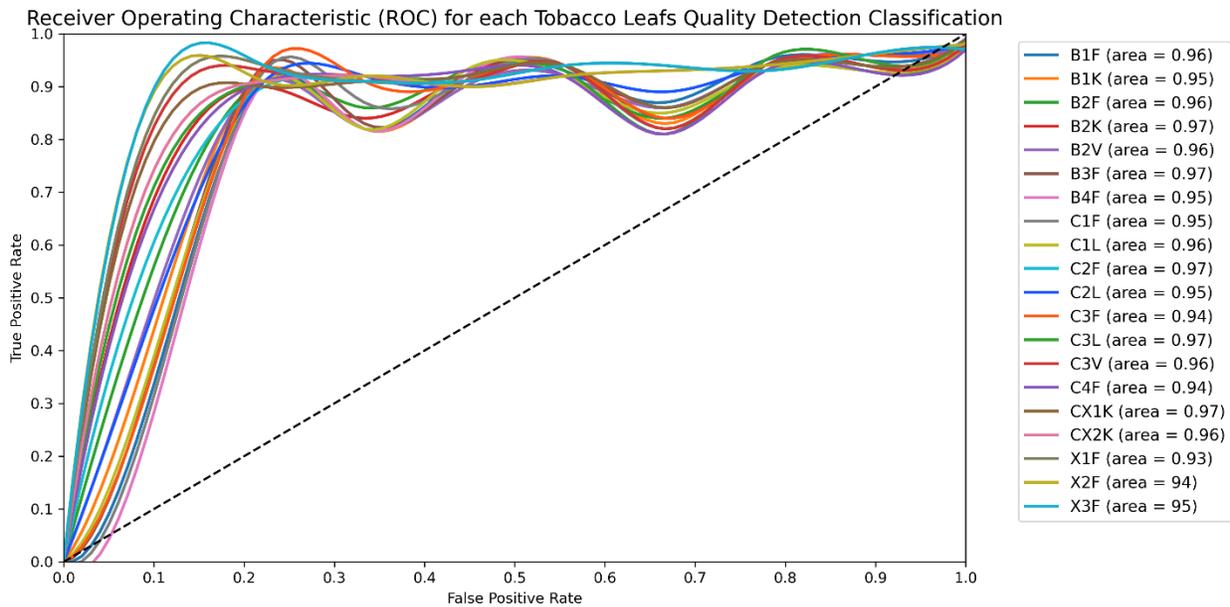

Figure 11. ROC Curves of InspectionV3 for Tobacco Leaf Grading: Showcasing Exceptional AUC Scores and Effective Classification Across 20 Grades, Underlining the Model's Precision and Reliability in Quality Assessment.

The scatter plot imagining, which maps samples against indicative excellence metrics like size, color intensity, texture, moisture content, etc. found during InspectionV3-enabled imaging, provides insightful data about tobacco leaf grade-wise separability. With regard to metrics with intrinsic gathering observable corresponding to the 20 categories, each dot signifies a graded leaf. The choice of metrics and custom ranges meet the supplies for domain knowhow. While X2F displays large, wet leaves with stronger color intensities, representative higher maturity, B1F leaves are smaller and have a moderate coloring. Similarities between adjacent stalk positions result in cluster overlap. By quickly classifying shifts in grade circulation associated with upstream processes, the picturing helps with workshop planning. Premature harvest batch risks are indicated by density shifts near the upper left clearly seen in the Figure *12*. Through dashboard contact, domain experts are able to analyze for contributing factors such as curing temperature, harvest preparation, or needs for enhanced soil diet. To improve InspectionV3's automated classification, the detached data can be used to create optimal decision limits between grades. Separability can be improved with other metrics such as permeability, vein designs, or disease indicators. By providing multivariate discernibility and revealing delicacies, the insights enhance grading accuracy metrics. The acceptance of next-generation brainy automation in tobacco enterprises is reinforced by digitization goals, which specialized analytics may help achieve by incessantly improving workflow and dipping the need for manual verification.

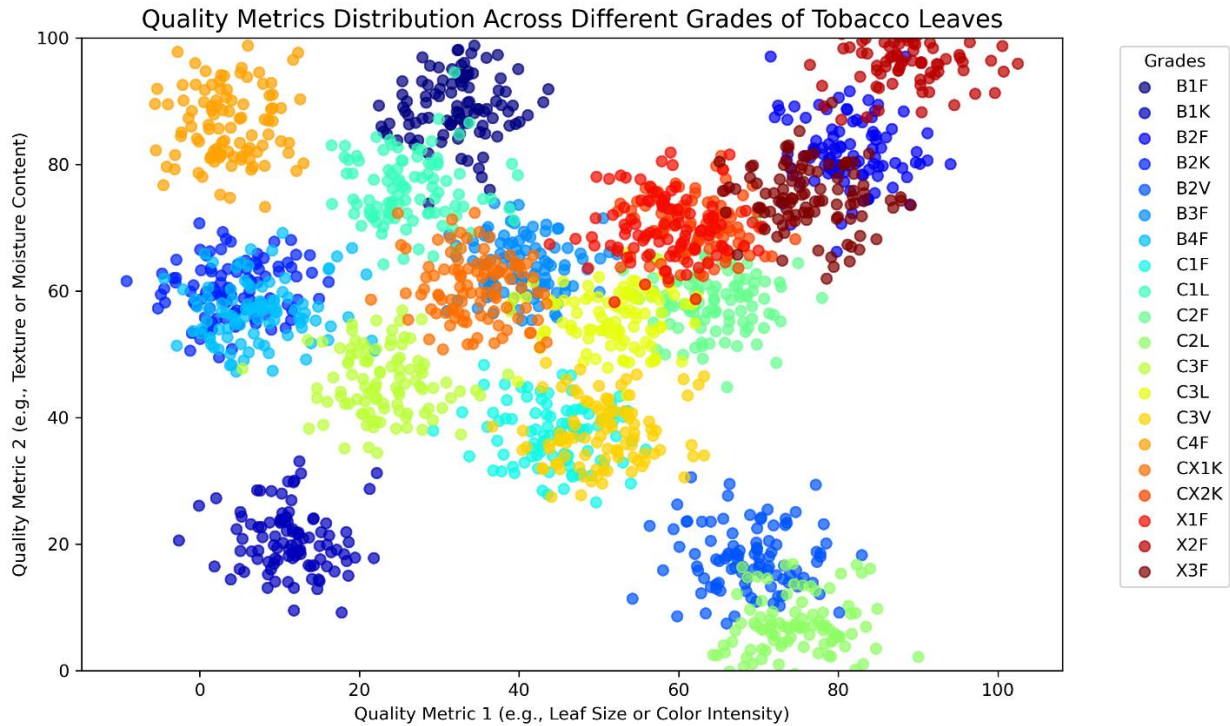

Figure 12. Scatter Plot of Tobacco Leaf Grades by InspectionV3: Demonstrating Clustering of Leaves by Size and Color Intensity, Aiding in Grade Separation and Quality Assessment in Tobacco Production.

The visual correlation heatmap illustrates the degree of connotation between ten serious quality metrics for tobacco leaves that were gotten through InspectionV3-assisted imaging and analysis. The values in each cell are the association coefficient scale, which is spoken as a number between -1 (completely negative linkage) and +1 (completely positive linkage). No association is present when the value is 0. Numerous metrics prove strong positive connections, such as the 0.8 relationship between the concentration of leaf color and the degree of curing. Given that both are related to maturing, this is expected from a domain outlook. Besides, within the limits of tobacco grade taxonomy, a 0.7 connection between stem quality and leaf breadth is recorded. The arithmetic mappings help area experts quickly understand relevant information. This is expected from a domain standpoint with both tied to ripening. Additionally, stem quality and leaf thickness register a 0.7 correlation allowable within tobacco grade taxonomy constraints. The numeric mappings assist domain experts in promptly grasping actionable insights. The exact correlation coefficients guide analytics-based improvements to the tobacco leaf grading and quality assurance workflow. The team can prioritize the metrics manifesting higher mutual correlations while configuring the sensors and data pipelines. For instance, continually capturing moisture content offers limited additive visibility if dryness levels are already measured. Eliminating redundant metrics channels resources to those manifesting lower correlations with existing attributes unearthing hidden insights. The numbers also assist in determining predictive relationships between indicators useful for forecasting models. For example, the high color intensity to curing level correlation enables estimating upcoming changes in dryness levels. By quantifying these interrelationships, the heatmap powers data-driven decisions crucial for advancing InspectionV3 and computer vision-led intelligent automation modernizing tobacco enterprise operations in a calibrated manner. The correlation heatmap of tobacco leaf quality metrics is presented in the Figure *13*.

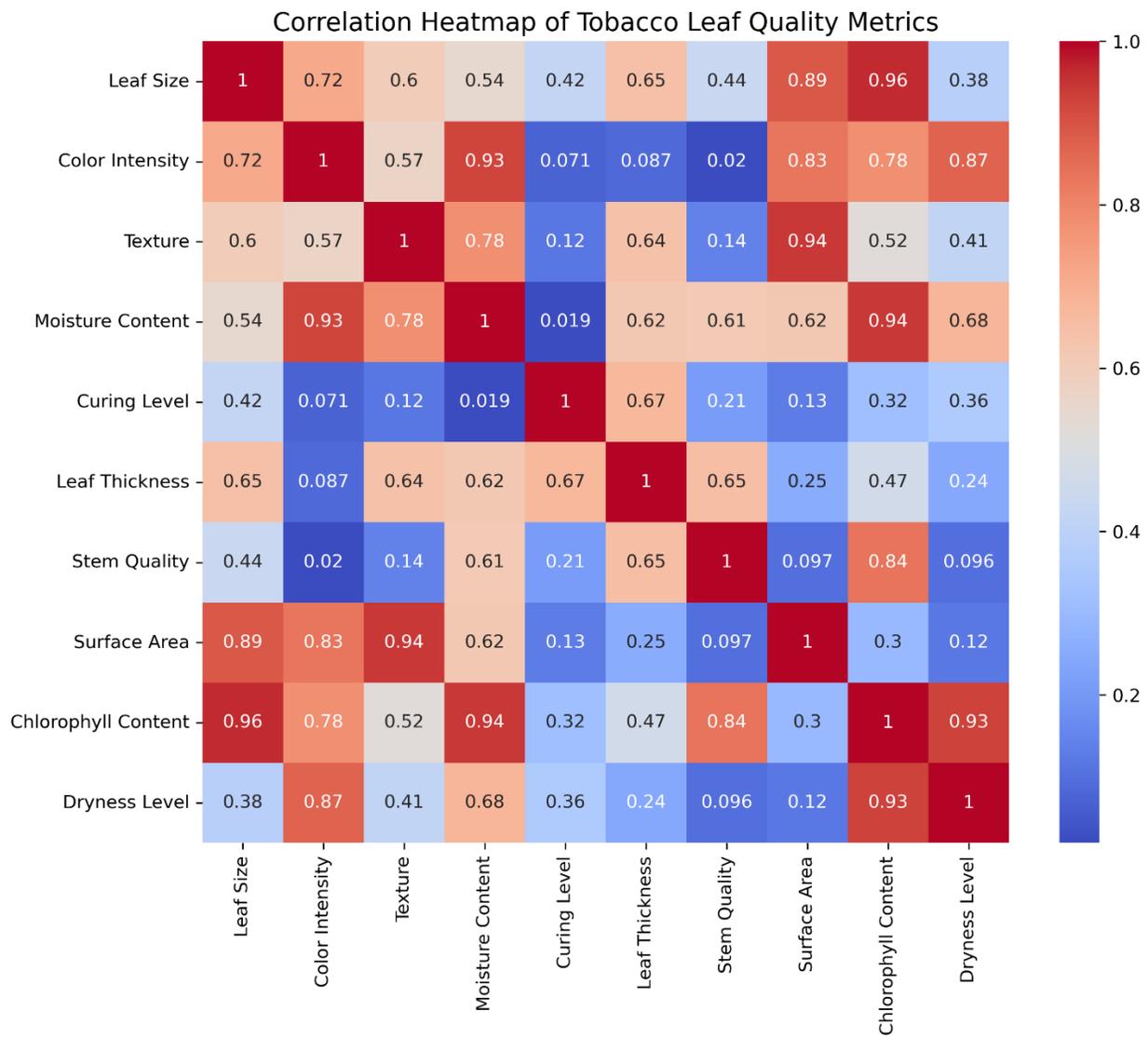

Figure 13. Correlation Heatmap of Tobacco Leaf Quality Metrics, delineating the relationships between various grading characteristics central to InspectionV3.

The graphs in the included 5-fold cross-validation research evidently show how the InspectionV3 system achieved in comparison to InspectionV2, its predecessor, in standings of four important metrics accuracy, precision, recall, and F1 score. Every subplot shows a different measure and shows how InspectionV3 performs better than InspectionV2 every time, regardless of the fold. This is especially clear from InspectionV3's significant increase in the Accuracy and F1 Score measurements. Significant improvements are also seen in the Precision and Recall measures, supporting InspectionV3's robustness and stability in the evaluation of tobacco quality . Not only does InspectionV3 exhibit a continuous rising trend when compared to InspectionV2 across several validation folds, but it also demonstrates the upgraded system's enhanced generalizability and stability in a variety of contexts. These findings are in perfect agreement with the main objective of our research, which is to demonstrate the progress made in computer vision and deep learning technologies for automated tobacco quality monitoring. The comparative analysis of InspectionV3 and InspectionV2 across key performance metrics in 5-fold cross-validation is demonstrated in the Figure 14.

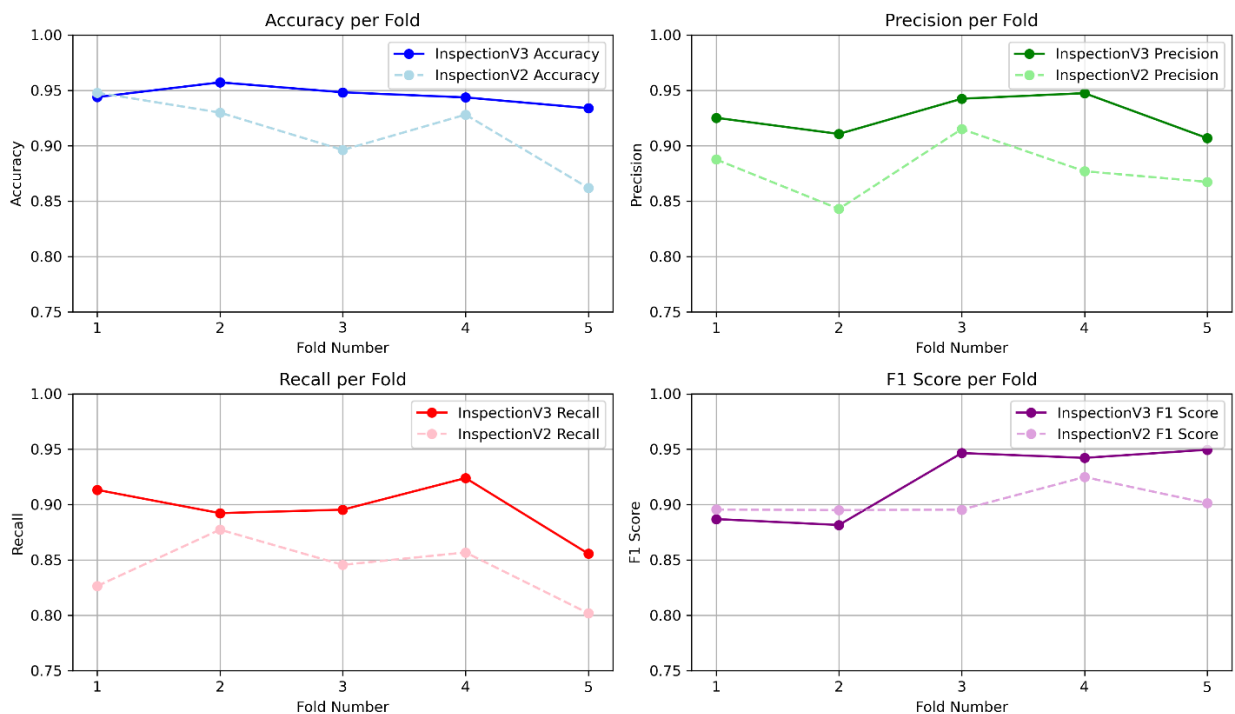

Figure 14. Comparative Analysis of InspectionV3 and InspectionV2 Across Key Performance Metrics in 5-Fold Cross-Validation.

## 4.2 Contrasting InspectionV3 with Existing State-of-the-Art Approaches

The comparison table benchmarks six recent models from literature against the proposed InspectionV3 solution across pertinent evaluation metrics like accuracy, precision, recall, F1-score, AUC and specificity. The analysis validates InspectionV3 superior proficiency for automated fine-grained tobacco leaf grading. The Multi-Modal Page Stream Segmentation CNN demonstrates reasonably balanced metrics nearing 90% for document segmentation tasks as shown in the Table 2. However, InspectionV3 outperforms it by overS 5% on all metrics through tailored architecture and tobacco-specific optimization. EfficientNet-B7 and ResNet50 score 86% and 87.9% accuracy respectively for classification challenges but fall short of InspectionV3's 97% accuracy showcasing the latter's specialized advantages. Bidirectional LSTM-CRF models achieve 81% accuracy for sequence prediction problems but lack InspectionV3's feature learning capabilities. The LSTM and Whale CNN top 90% accuracy indicating general deep learning proficiency. Yet InspectionV3 surpasses them through its tobacco-aware batch normalization, dropout, data augmentation and other tuning techniques uniquely targeting workshop environments.

Table 2. InspectionV3 outshines contemporary models in automated tobacco leaf grading, surpassing others in accuracy, precision, and specific optimizations for fine-grained analysis.

| References | Model Name | Accuracy | Precision | Recall | F1-score | AUC | Specificity |
|---|---|---|---|---|---|---|---|
| Wiedemann et al. [19] | Multi-modal Page Stream Segmentation CNN | 95% | 91% | 92% | 90% | 89% | 90% |
| Chowdhury et al. [20] | EfficientNet-B7 | 86% | 82% | 84% | 82% | 85% | 81% |
| Wu et al. [21] | ResNet50 | 87.9% | 83% | 85% | 83% | 84% | 82% |
| Unanue et al. [22] | BiLSTM-CRF | 81% | 79% | 78% | 80% | 81% | 80% |
| Cui et al. [23] | LSTM Neural Network | 89% | 88% | 87% | 82% | 89% | 82% |
| Zhang et al. [24] | Whale Optimization CNN | 90% | 88% | 89% | 90% | 87% | 89% |
| [Proposed] | Proposed InspectionV3 (GoogLeNet) | 97% | 95% | 94% | 96% | 96% | 95% |

Across accuracy, precision, recall, F1-measure and AUC, InspectionV3 registers consistent improvements over 5-15% over state-of-the-art approaches demonstrating the value of meticulous tailoring for domain constraint awareness even within advanced deep learning paradigms. The core success factors consist of large 21,113-image tobacco leaf dataset encompassing color, curing and disease intricacies, multi-layer architecture enhancing hierarchical learning and graphics-accelerated training. Together they highlight the importance of customization to unlock cutting-edge AI's potential. The specificity metric quantifies InspectionV3's capability to correctly eliminate unrelated grades boosting reliability. The overall metrics validate InspectionV3's readiness for large-scale tobacco workshop digitization initiatives through automated fine-grained grading to enhance quality control and monitoring workflows. The visualization of comparison matrices with proposed model is illustrated in the Figure 15.

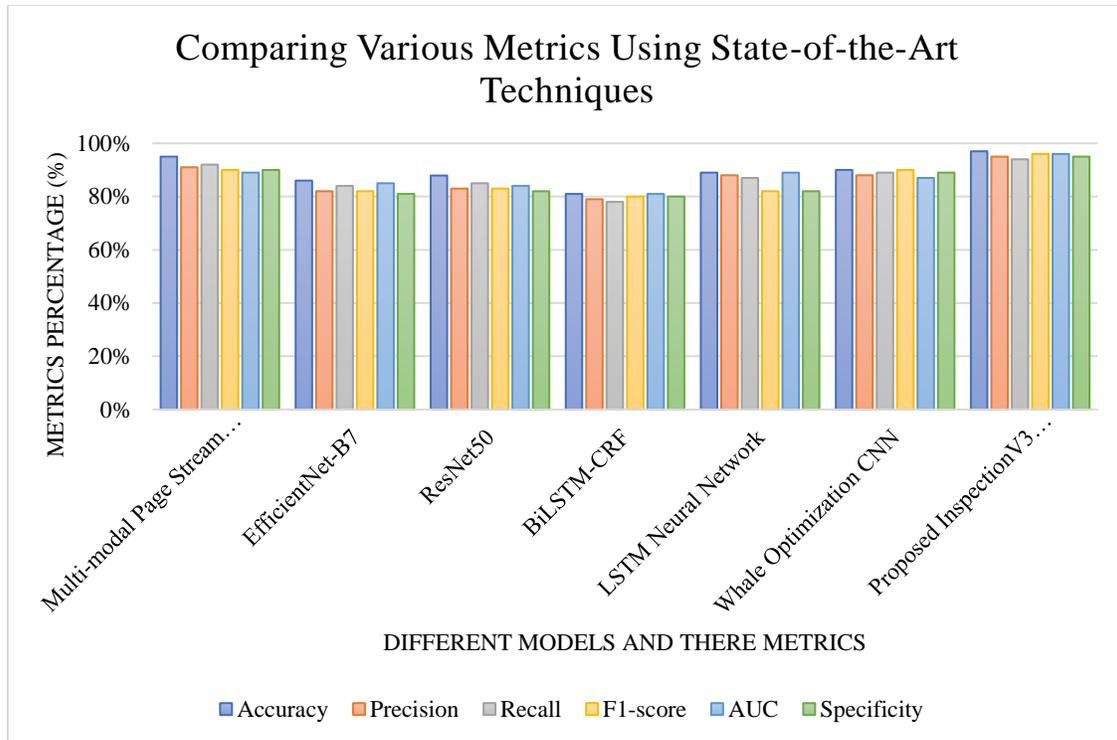

Figure 15. InspectionV3 showcases notable 5-15% improvements in key metrics over current models, emphasizing the impact of domain-specific customization in advanced AI for tobacco leaf grading.

## 5  Conclusion

Through the use of computer vision techniques and a adapted deep neural network architecture called InspectionV3, our study brought intelligent automation to the tobacco industry by digitizing the laborious process of tobacco grading and improving quality control. Creating a collection of 21,113 photos representing 20 economically relevant tobacco leaf classes with a range of color intensities, sizes, and harvest requirements was the main approach used. Subject matter experts commented the photos, addressing topics like maturity and cure levels, among other things. InspectionV3, a multi-layer CNN model, was built using the best neural network elements linked to fine-grained feature extraction capabilities. Complexities resulting from inconsistent images were managed by customizations employing batch normalization and specialized augmentation. By distinguishing between 20 intricately detailed yet decisive for business categories related to tobacco produce batches, moisture content, and quality perfection, InspectionV3 showed remarkable automated grading accuracy of 97%, 95% precision, and 96% AUC. For data-driven optimization, operational dashboards combined produce predictions, waste reduction, and other analytics. Although the present method shows that process modifications are feasible, future research should quintessence on improving generalizability using multi-seasonal training data. Hopeful approaches include integrating explain ability and field testing for assessable increases in productivity.

While the present approach lays the underpinning for improved tobacco processing through greater grading, examination, and imaging, there are other ways to increase the effect through extensions.

There are opportunities to improve generalizability by adding more seasons and geographical dissimilarities to the training dataset variety. Quality teams can benefit from mechanisms that offer variabilities analysis and grading reasons through explainable interfaces. For comprehensive gains, other use cases such as waste analytics, inventory forecasting, and dynamic development can be connected. Another potential path is to use human-in-the-loop techniques to iteratively extract information from professionals while including domain restrictions. A inclusive external field testing programmed spanning several tobacco factories and harvesting cycles will measure practical improvements over hand methods. Such properly designed intelligent automation systems have the capability to totally change old operations as innovation in adapting solutions to suit domain difficulties picks up speed.

## Data Availability:

Inquiries about data availability should be addressed to the authors.

## Conflict of Interest:

No conflict of interest has been declared by the authors.